\pdfoutput=1
\documentclass[11pt]{article}

\usepackage{acl}

\usepackage{times}
\usepackage{latexsym}
\usepackage{amsmath}
\usepackage{graphicx}
\usepackage{natbib}
\usepackage{amsfonts}
\usepackage{multirow}
\usepackage{amssymb}
\usepackage{arydshln}

\usepackage[T1]{fontenc}
\usepackage[utf8]{inputenc}

\usepackage{microtype}

\title{Time Machine GPT}

\author{Felix Drinkall*, Eghbal Rahimikia §, Janet B. Pierrehumbert*\ddag, Stefan Zohren*\dag \\
    *Department of Engineering Science, University of Oxford \\
    § Alliance Manchester Business School, University of Manchester \\
    \dag The Alan Turing Institute \\
    \ddag Faculty of Linguistics, University of Oxford \\
    \texttt{felix.drinkall@eng.ox.ac.uk}}

\begin{document}
\maketitle
\begin{abstract}
Large language models (LLMs) are often trained on extensive, temporally indiscriminate text corpora, reflecting the lack of datasets with temporal metadata. This approach is not aligned with the evolving nature of language. Conventional methods for creating temporally adapted language models often depend on further pre-training static models on time-specific data. This paper presents a new approach: a series of point-in-time LLMs called \textbf{Ti}me\textbf{Ma}chine\textbf{GPT} (TiMaGPT), specifically designed to be nonprognosticative. This ensures they remain uninformed about future factual information and linguistic changes. This strategy is beneficial for understanding language evolution and is of critical importance when applying models in dynamic contexts, such as time-series forecasting, where foresight of future information can prove problematic. We provide access to both the models and training datasets.\footnote{Models and Datasets: https://huggingface.co/Ti-Ma}
\end{abstract}

\section{Introduction}

\begin{figure}[t!]
    \centering
    \includegraphics[width=1\linewidth]{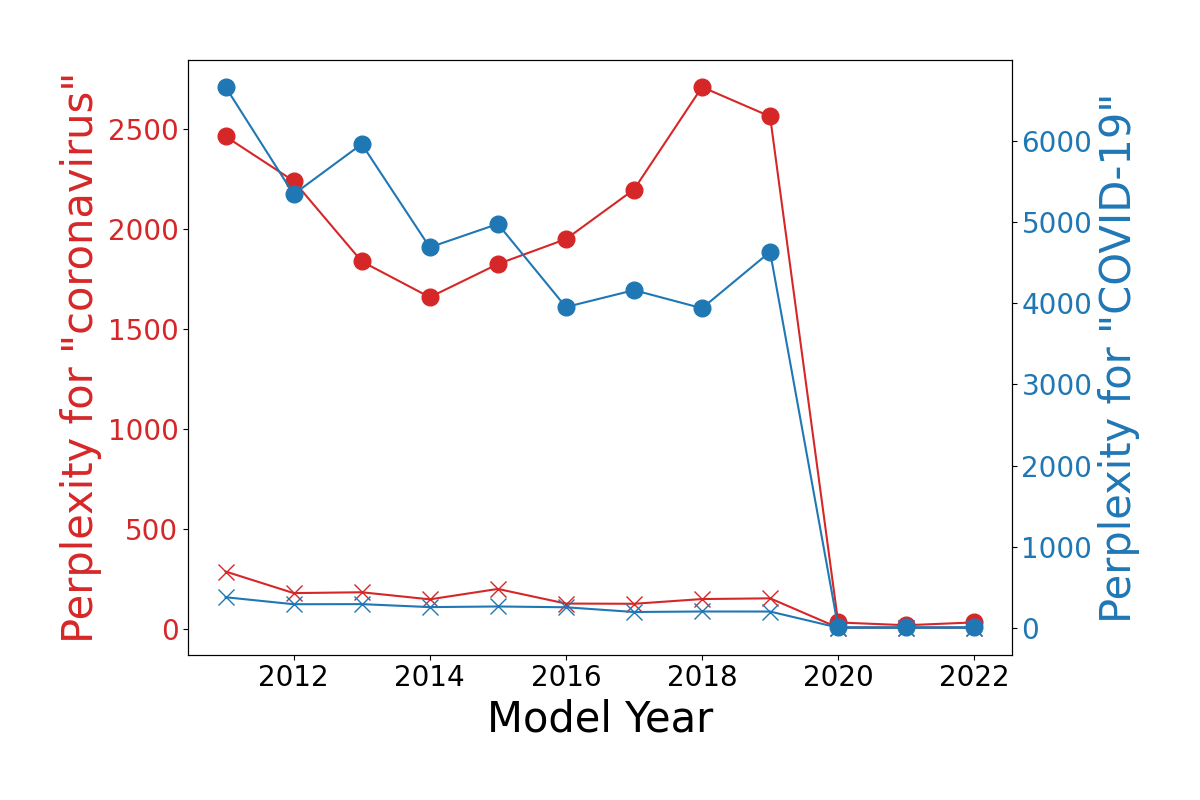}
    \caption{The perplexity of \textcolor{red}{coronavirus} and \textcolor{blue}{COVID-19}, using TiMaGPT models (\textbullet{}) and Conventional Temporally Adapted (CTA) models ($\times$). The calculation for perplexity is outlined in Appendix \ref{app:perp} and the methodology for temporally adapting models is explained in Section \ref{sec:eval}. The CTA models have significant knowledge of these words before the pandemic.}
    \label{fig:covid_perp}
\end{figure}

Time-series forecasting and event prediction aim to infer a future state of the world from past data. When evaluating models for these purposes through historical data analysis, often referred to as "back-testing", it is crucial to maintain strict data partitioning. This ensures that no future information influences the model's predictions. Whilst strict data partitioning is standard in most fields that use time-series information, time-series forecasting methods that use transformer-based LLMs have tended to make an assumption that the language model itself cannot be the vector for information leakage from a future state to a past state. However, within a language model, implicit associations, such as linking "Enron" with "bankrupt" or possessing knowledge of terms like "COVID-19" might exist (Figure \ref{fig:covid_perp}). This poses a challenge for models tested on data predating such events, as their presence could lead to an overestimate in model performance within the validation stage, which could lead to disappointing results when a system is used in a live setting.

The evolution of language models in recent years has been shaped by increases in both the size of these models and their training datasets \cite{wei2022emergent}. The trend towards larger and more complex datasets has made in-depth analysis of their content increasingly difficult. A significant challenge is the contamination of training datasets, which can include the accidental inclusion of benchmark datasets \cite{dodge-etal-2021-documenting} and private data \cite{xu-etal-2020-personal}. To tackle the issue of temporal data contamination, this paper introduces language models that have been pre-trained on data exclusively published before specified cutoff dates. These models serve two key purposes: analyzing diachronic embeddings over time and facilitating the use of language models in dynamic tasks that demand strict separation of temporal data. Language models are capable of learning both factual information and linguistic patterns \cite{petroni-etal-2019-language, mahowald2023dissociating}, which could influence their performance in predictive tasks. The primary application for our TiMaGPT models is in evaluating a system that uses generative language models for dynamic downstream tasks. The yearly models, developed uniformly, display similar performances on well-established benchmarks, meaning that the main difference between the models is the information in the training datasets.

\subsection{Research Contributions}

\textbf{Contribution 1} - To our knowledge, the models released in conjunction with this paper are the first series of temporally correct pre-trained LLMs exclusively pre-trained on historical data. 
\\
\textbf{Contribution 2} - Identification of an unacceptable level of foresight in conventional temporally adapted models.

\section{Related Work}

\subsection{Diachronic Embeddings}

The meaning of words change subject to the context in which they appear, a fact that initiated the adoption of contextualized embeddings over static embeddings within LLMs \citep{mikolov2013efficient, NIPS2013_9aa42b31, devlin-etal-2019-bert}.  This contextual dependency extends beyond the surrounding tokens, as words can change meaning according to the venue \cite{zeng2018biased}, domain \cite{10.1093/bioinformatics/btz682, yang2020finbert}, time \citep{pierrehumbert2012dynamic, bybee_2015}, location \citep{dunn-2023-variation, hofmann2023geographic}, or task \citep{gururangan-etal-2020-dont}. 

Consequently, there has been significant research focused on understanding how embeddings shift through time. Procrustean alignments enabled \citet{kutuzov-etal-2017-tracing} to assess the way in which word meaning shifted by diachronically training static embeddings. Some have tried to incorporate these temporal dynamics into LLMs by dynamically adapting word embeddings \citep{rudolph2017dynamic, hofmann-etal-2021-dynamic}. Numerous studies have investigated how these embeddings evolve over time \citep{hamilton-etal-2016-diachronic, kutuzov-etal-2018-diachronic}, with practical applications such as detecting change points in language use \citep{goutte-etal-2018-eurogames16}. These studies demonstrate that embeddings can reveal the temporal context of data, underscoring the importance of carefully selecting the data included in training datasets.

\subsection{Temporal Adaptation of Language Models}

Efforts to temporally adapt language models to date have primarily involved modifying existing statically trained models \cite{lazaridou2021mind, rottger-pierrehumbert-2021-temporal-adaptation, dhingra-etal-2022-time}. Given that transformer-based LLMs have been predominantly trained since 2017, following the seminal work of \citet{vaswani2023attention}, and largely on data from post-2017, temporal adaptation has generally involved either further training these models with newer data \cite{jang2022continual} or adjusting them to represent a past state by training on historical data for a fixed number of steps (hereafter "CTA models" - \textbf{C}onventional \textbf{T}emporally \textbf{A}dapted) \cite{https://doi.org/10.13140/rg.2.2.14905.44649, martinc-etal-2020-leveraging}. Both methods have significant limitations, since either any resultant downstream analysis is limited to the very short time after the models were trained, or the temporally adapted models have seen future data within the pre-training stage. This paper restores language models to a prior state in time by pre-training a series of models on data that has strict temporal inclusion criteria.

\section{Training Process}

\subsection{Training Datasets}
\label{sec:datasets}

The lack of temporal metadata in natural language processing (NLP) presented a challenge in selecting datasets for training our models. However, news data and Wikipedia version history emerged as valuable resources. Detailed token counts for each year's deduplicated datasets are provided in Appendix \ref{app:tok_counts}. Each year from 2011 to 2022 contained sufficient data to train a GPT-2 small model.

\textbf{Wikipedia}: By utilizing the revision information from Wikipedia XML dumps provided by Wikimedia\footnote{https://dumps.wikimedia.org}, we reconstructed every existing Wikipedia page as they would have appeared on 31/12 of each year from 2004 to 2023. This reconstruction accounted for changes in page titles. The identified revisions were then processed to remove links, HTML, and other non-standard stylistic elements, using the following code repository \footnote{http://tinyurl.com/2exawtkf}.

\textbf{WMT News}: The WMT News dataset, typically used in machine translation \cite{kocmi-etal-2022-findings}, was processed in its monolingual, document-split English version. We applied deduplication to this dataset, eliminating repeated articles via an SHA-256 hashing function \cite{chenghao_mou_2023_8364980}. The dataset ranges from 2007 to 2022.

\subsection{Dataset Aggregation}
\label{sec:data_agg}

Several studies have demonstrated that the data types used in training an LLM significantly influence its performance in downstream tasks. This insight led to the development of domain-specific language models such as BioBERT \cite{Lee_2019}, SciBERT \cite{beltagy-etal-2019-scibert}, FinBERT \cite{yang2020finbert}, and more recent models like BloombergGPT \cite{wu2023bloomberggpt}. Acknowledging this, we maintained a consistent token allocation from each domain in our annual datasets. This approach ensured that the language models' performance wasn't skewed by shifts in the relative size of different data domains over time. Consequently, the only differences among the various training datasets are the new information and time-specific stylistic changes unique to each period.

\subsubsection{Sampling}
\label{sec:sampling}

To maintain a predetermined domain allocation ratio of 0.6:0.4 (WMT News to Wikipedia), a ratio that was determined by model tuning outlined in Appendix \ref{app:tuning}, we employed specific sampling strategies for each dataset.

We randomly sampled Wikipedia articles from each year, ensuring articles were not chosen twice. Additionally, we included the "Vital Level 4" pages – the top 10,000 most important Wikipedia articles \footnote{Level 4 Vital Articles: https://tinyurl.com/532uaexs} – in each training dataset. The Level 4 articles changed slightly over time, so our selection was based on the list available at the end of each year.

For the WMT News dataset, ordered as a text stream, we have included data according to a negative exponential probability function over a 5-year period to prioritize recent data over older data. We first identify the start date of the 5-year window and calculate the number of days from the cutoff date, represented as $\tau$. For each entry $e_i$ with an age of $D_i$ days in the dataset, we compute a weight that is assigned to each entry based on its age, given by:
\begin{equation}
    W_i = \exp\left(-\frac{D_{\text{max}} - D_i}{\tau}\right)
\end{equation}

The probability of selecting each entry, $P_i$, is inversely proportional to its weight, such that:
\begin{equation}
    P_i = \frac{1/W_i}{\sum_{j=1}^{N} \frac{1}{W_j}}
\end{equation}
where $N$ is the number of entries in the dataset.

In the process of sampling, our goal is to accumulate a certain number of tokens, denoted as $T_{\text{needed}}$. Starting with an initial token count of $T_{\text{current}} = 0$, we repeatedly sample  with probability $P_i$ until:
\begin{equation}
    T_{\text{current}} \geq T_{\text{needed}}
\end{equation}

If adding the tokens of a chosen entry does not exceed $T_{\text{needed}}$, we add the entry to the training dataset and update the token count $T_{\text{current}}$.

\begin{table*}[h]
    \centering
    \scalebox{0.95}{
    \begin{tabular}{|c||c||c|c|c|c|c|}
        \hline
        \multirow{2}{*}{Model} & \multicolumn{6}{c|}{Benchmark Performance} \\
        \cline{2-7}
         & Av. & HellaSwag & PIQA & TruthfulQA & Winogrande & WSC \\
         \hline 
        Baseline & 39.5 & 25 & 50 & 22.5 & 50 & 50 \\
        \hline
        GPT-2 Small & 45.85 & 31.14 & 62.51 & 40.69 & 51.62 & 43.27 \\
        OPT 125m & 44.60 & 31.34 & 62.02 & 42.87 & 50.20 & 36.54 \\
        GPT-Neo 125m & 45.08 & 30.26 & 62.46 & 45.58 & 50.43 & 36.54 \\
        \hline
        TiMaGPT$_{'11}$ & 48.74 & 25.14 & 50.87 & 52.83 & 51.38 & 63.46 \\
        TiMaGPT$_{'12}$ & 48.69 & 25.26 & 50.98 & 53.30 & 50.99 & 63.46 \\
        TiMaGPT$_{'13}$ & 48.62 & 25.12 & 50.82 & 53.11 & 50.36 & 63.46 \\
        TiMaGPT$_{'14}$ & 48.61 & 25.04 & 50.27 & 52.88 & 50.04 & 63.46 \\
        TiMaGPT$_{'15}$ & 48.75 & 24.98 & 50.76 & 52.74 & 50.59 & 63.46 \\
        TiMaGPT$_{'16}$ & 48.99 & 25.00 & 50.27 & 52.60 & 51.62 & 63.46 \\
        TiMaGPT$_{'17}$ & 48.98 & 25.09 & 50.76 & 52.25 & 51.62 & 63.46 \\
        TiMaGPT$_{'18}$ & 48.43 & 25.13 & 51.31 & 52.41 & 49.64 & 63.46 \\
        TiMaGPT$_{'19}$ & 48.66 & 25.30 & 50.98 & 52.30 & 50.83 & 63.46 \\
        TiMaGPT$_{'20}$ & 48.65 & 25.07 & 50.77 & 52.88 & 51.14 & 63.46 \\
        TiMaGPT$_{'21}$ & 48.58 & 25.38 & 51.52 & 52.55 & 50.67 & 63.46 \\
        TiMaGPT$_{'22}$ & 48.52 & 25.34 & 51.47 & 52.90 & 50.04 & 63.46 \\
        \hline
    \end{tabular}
}
    \caption{Performance of the models on static benchmarks to validate performance. HellaSwag, TruthfulQA, PIQA, Winogrande, WSC (Appendix \ref{app:bench_datasets}). Comparison models: GPT-2 \cite{radford2019language}, OPT \cite{zhang2022opt}, GPT-Neo \cite{gpt-neo}}
    \label{table:model_comparison}
\end{table*}

\subsection{Pre-training Details}

The full training details for replicating our work are provided in Appendix \ref{app:training_details}. In line with the Chinchilla ratio, which recommends a 1:20 parameter-to-token ratio for efficient training \cite{hoffmann2022training}, a GPT-2 model with 117 million parameters requires 2.34 billion tokens for optimal training. We trained each of our models on 2.5 billion tokens and used a BPE tokenizer as was used in the original GPT-2 paper \citet{radford2019language}. To confirm that this amount of data was sufficient, we performed a comparative analysis of models trained with varying token counts, detailed in Appendix \ref{app:tok_counts}. Considering the numerous models we had to train, we optimized our training framework for computational efficiency. Therefore when two samples could be combined into the 1024 token sequence we concatenated them. A similar methodology only saw a marginal reduction in performance when training RoBERTa \cite{liu2019roberta}.

\section{Model Verification}

Verifying that each of our models achieves an adequate level of performance is essential. To conduct meaningful analysis on downstream tasks, it is vital to ensure consistent performance on static benchmarks from models from different years. This consistency means that we can assume that the majority of any observed changes are due to variations in the information within the training datasets, not fluctuations in model efficacy. When selecting candidate benchmarks, we observed that for some newer and more complex benchmarks models of this size have a performance similar to the random baseline. This is due to the rapid progress in language model performance in recent years, and the need to create new benchmarks to match that progress. A more detailed description of the tasks that were included is in Appendix \ref{app:bench_datasets}. Table \ref{table:model_comparison} demonstrates that while our models are far from the state-of-the-art, they perform in line with other similarly-sized models like GPT-2, OPT-125m and GPT-Neo 125m on several established benchmarks but crucially also maintain this performance over time. TiMaGPT has a slightly different performance profile to the comparison models, with better performance on the WSC and TruthfulQA benchmarks and worse performance on the Hellaswag and PIQA datasets. Interestingly, both of the benchmarks that TiMaGPT performed badly on were challenging commonsense reasoning datasets. Perhaps the factual bias and lack of diversity of our training training data led to poor performance on these benchmarks. All of the models perform just slightly above random for the Winogrande benchmark, indicating that this benchmark is too challenging for models of this type and size. The lack of variance of the TiMaGPT results on the WSC benchmark can be attributed to the dataset's size - only 273 samples in total. 

\section{Temporal Evaluation}
\label{sec:eval}

Previously, models were adapted by further training a statically trained model on period-specific data \cite{histBERT, dhingra-etal-2022-time}, giving them foresight from the pre-training stage, which could be problematic for tasks where temporal segregation is important. We compared our models with Conventionally Temporally Adapted (CTA) models to show the extent of the information leakage when adopting the traditional methodology, by assessing their perplexity in recognizing the names of country leaders around their inauguration. The perplexity measurement is outlined in Appendix \ref{app:perp} and the dataset identifies leaders that came into power between 2013 and 2020 \cite{herre_2023}. 310 leaders are considered, corresponding to 154 countries. 

\begin{figure}[b!]
    \centering
    \includegraphics[width=1\linewidth]{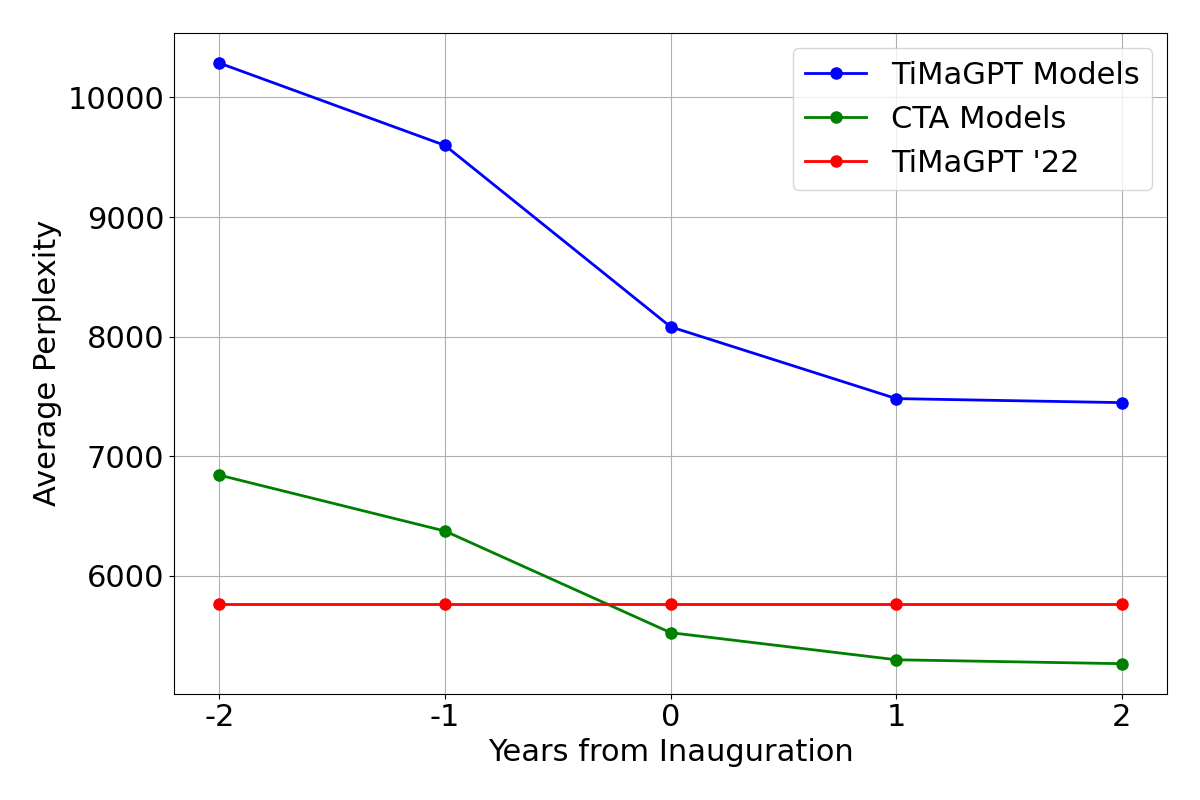}
    \caption{Average perplexity of the names of country leaders around their year of inauguration, as measured using CTA models (Section 2.2) and TiMaGPT models.}
    \label{fig:perp_comp}
\end{figure}

We contrasted our TiMaGPT models with CTA models, which are versions of the TiMaGPT$_{2022}$ model further pre-trained on 1 billion tokens from the same datasets used for pre-training the yearly TiMaGPT models. Figure \ref{fig:perp_comp} shows the differences in methodologies, with CTA models retaining unrealistic knowledge of the leaders well in advance of their inauguration. The lower CTA perplexity scores come from seeing the leaders in the pre-training dataset that trained TiMaGPT$_{2022}$, providing the CTA models with information that would not have been available to models that were trained at the time. The relatively low perplexity from TiMaGPT$_{2022}$ supports this claim. It is clear from the performance delta between the CTA models and the TiMaGPT$_{2022}$ model two years before inauguration that whilst temporal adaptation does shift the model's perplexity distribution closer to that of the TiMaGPT models, some information is preserved post-adaptation.

Beyond named entities, we show that CTA models have an unrealistic knowledge of concepts like "COVID-19" or "coronavirus"; Figure \ref{fig:covid_perp} exposes a very significant difference between our models and traditional adaptation methods. The CTA model perplexity scores are lower than our TiMaGPT models, which reflect what could have been produced at the time. The TiMaGPT dataset partitioning means that the information leakage seen in Figure \ref{fig:covid_perp} and \ref{fig:perp_comp} does not occur.

\section{Discussion}
\label{sec:disc}

This paper provides a tool for researchers focused on tracking knowledge and association shifts in language, and also in evaluating the performance of temporally dynamic models. The recent trend of using GPT-2 as a backbone for time-series forecasting, as highlighted in recent literature \cite{cao2023tempo, chang2023llm4ts, zhou2023fits, liu2024unitime}, underscores the growing interest in integrating language models and textual features to enhance forecast accuracy \cite{drinkall-etal-2022-forecasting, cao2023tempo}. The models developed in conjunction with this paper are particularly valuable in this context. They serve as an effective means to minimize look-ahead bias in time-series models that concurrently process textual and time-series data. By ensuring these models are devoid of future linguistic information, they enable a more accurate and authentic assessment of a model's forecasting ability, crucial for applications where current data must be interpreted without the influence of future events. Further work could explore the magnitude of the effect of this look-ahead bias by measuring the performance delta between models that have and have not seen future information in their pre-training.

\section{Limitations}
\label{sec:limitations}

The paper uses the small GPT-2 architecture, which is outperformed by many newer language models. To create larger TiMa models, it is necessary to expand the size and number of datasets with temporal metadata. This expansion is crucial because each parameter in these models requires around 20 tokens for optimal pre-training \cite{hoffmann2022training}. In addition, we have only explored generative models in this paper, but a significant amount of research still relies on encoder-based LLMs which limits the scope of this paper.

To scale to even larger models, processing the annual Common Crawl datasets is a necessary step, though the dataset has proved problematic due to its scale and lack of consistent formatting \cite{luccioni-viviano-2021-whats}. These problems prompted the C4 dataset \cite{dodge-etal-2021-documenting}, but replicating that consistent quality over several partitioned years would be a significant challenge. Aside from this, cleaning Common Crawl would also demand significant computational resources.

\section*{Acknowledgements}

We are grateful to Graphcore, and their team, for their support in providing us with compute for this project. The first author was funded by the Economic and Social Research Council of the UK via the Grand Union DTP. This work was supported in part by a grant from the Engineering and Physical Sciences Research Council (EP/T023333/1). We are also grateful to the Oxford-Man Institute of Quantitative Finance and the Oxford e-Research Centre for their support. 

\bibliography{anthology,custom}
\bibliographystyle{acl_natbib}
\appendix
\begin{table*}[h!]
    \centering
    \begin{tabular}{c|ccc|cc}
        Year & WMT & WMT Cumulative & WMT 5-MS & Wiki & Wiki Core \\
        \hline
        2007 & 115,072,991 & 115,072,991 & \textcolor{red}{115,072,991} & 2,683,520,653 & 11,867,169 \\
        2008 & 413,793,002 & 528,865,993 & \textcolor{red}{528,865,993} & 3,736,056,257 & 19,044,576 \\
        2009 & 504,632,842 & 1,033,498,835 & \textcolor{red}{1,033,498,835} & 4,581,675,532 & 27,362,228 \\
        2010 & 233,111,988 & 1,266,610,823 & \textcolor{red}{1,266,610,823} & 5,311,904,669 & 35,398,949 \\
        \hdashline
        2011 & 505,374,950 & 1,771,985,773 & 1,771,985,773 & 6,146,126,877 & 78,283,040 \\
        2012 & 427,188,977 & 2,199,174,750 & 2,084,101,759 & 6,782,268,690 & 88,187,713 \\
        2013 & 727,323,818 & 2,926,498,568 & 2,397,632,575 & 7,105,210,758 & 86,770,551 \\
        2014 & 724,859,204 & 3,651,357,772 & 2,617,858,937 & 7,662,142,757 & 94,680,128 \\
        2015 & 725,113,377 & 4,376,471,149 & 3,109,860,326 & 8,407,835,670 & 95,613,538 \\
        2016 & 558,931,038 & 4,935,402,187 & 3,163,416,414 & 8,801,952,709 & 97,948,198 \\
        2017 & 928,705,556 & 5,864,107,743 & 3,664,932,993 & 9,449,623,447 & 103,278,211 \\
        2018 & 559,133,658 & 6,423,241,401 & 3,496,742,833 & 9,699,735,445 & 76,140,734 \\
        2019 & 799,069,641 & 7,222,311,042 & 3,570,953,270 & 9,868,604,683 & 71,284,359 \\
        2020 & 1,049,834,674 & 8,272,145,716 & 3,895,674,567 & 10,105,269,307 & 90,346,479 \\
        2021 & 1,016,847,474 & 9,288,993,190 & 4,353,591,003 & 10,208,296,406 & 74,019,900 \\
        2022 & 1,067,806,539 & 10,356,799,729 & 4,492,691,986 & 8,543,710,700 & 73,433,918 \\
    \end{tabular}
    \caption{Token counts of the base domain datasets after cleaning and deduplication; WMT: the token count of the articles from that year; WMT Cumulative: represents the token count of all WMT articles before each cut-off date; WMT 5-MS: the moving sum of the preceding 5 years of WMT data, which is all the data that we sample from for each year; Wiki: the token count from the whole Wikipedia yearly partition; Wiki Core: the token count of the Level 4 Vital Wikipedia pages.}
    \label{tab:token_count}
\end{table*}

\section{Risks}
\label{sec:risks}

The risks associated with this paper are not that significant due to the type of data used. The datasets and benchmarks in this paper are all open source. Long-term use of the WMT News datasets minimizes the chance of persisting errors. However, Wikipedia data, editable by anyone, could be less reliable. The selected December 31st revision may have inaccuracies. We haven't taken additional measures to verify the truthfulness of the content. A study by \citet{Kraenbring2014AccuracyAC} found Wikipedia's pharmacology information 99.7\% accurate, but this may not hold true for other subjects.

\section{Token counts}
\label{app:tok_counts}

The base datasets grow and shrink over time. Our sampling method from Section \ref{sec:sampling} means that the domain split of the data stays static across our models. Table \ref{tab:token_count} tabulates the overall token counts of the cleaned deduplicated datasets from which the training datasets are made. 

\section{Training details}
\label{app:training_details}

All models are trained on a Graphcore IPU-POD16 using the gpt2-small-ipu config, which employs tensor sharding for efficient distribution across multiple IPUs. We use the AdamW optimizer with
\( \beta_1 = 0.9 \), \( \beta_2 = 0.95 \), \( \epsilon = 10^{-8} \),
and a weight decay of 0.1. We adopt a linear warm up from $0.1*LR_{max}$ to $LR_{max}=31*10^{-5}$ over 10 percent of the training data. The subsequent learning rate was determined by a linear scheduler from $LR_{max}$ to $0.1*LR_{max}$ over the rest of the training data.

The models were designed with a context span of 1024 and configured to generate sequences of up to 50 tokens. We adopted the GPT2LMHeadModel with the GELU new activation function, comprised of 12 layers and 12 attention heads, and an embedding dimension of 768. A single seed was used to initialise the training of all of the models.

\section{Wikipedia Processing}
\label{app:wiki}

In conjunction with this paper, we are releasing the yearly partitions of Wikipedia that were instrumental in creating our training datasets \footnote{https://huggingface.co/Ti-Ma}. WikiMedia routinely publishes dumps of Wikipedia, each containing the revision history of articles. With approximately 60 million articles on Wikipedia, many having thousands of revisions, processing these revisions demands substantial computational resources. To streamline this process, we first defined the relevant revision before extracting the article information. Specifically, we select the most recent revision as of December 31st for each year. Consequently, some revisions in our datasets, such as those in the 2020 training set, date back to before 2006, as illustrated in Figure \ref{fig:wiki_hist}. While this inclusion of older revisions might initially appear problematic, it is important to note that these are the existing versions of Wikipedia pages as of the cutoff date. The content of these pages was considered current enough at that time, implying that a more recent revision was not necessary. This approach ensures that our training datasets reflect the most up-to-date information available on Wikipedia at each year's end, providing a realistic snapshot of knowledge for that specific point in time. 

Once each revision has been identified we clean the page using the code from \textit{wiki-dump-reader} \footnote{https://github.com/CyberZHG/wiki-dump-reader/tree/master}, which parses the page and outputs clean text. During the cleaning phase a number of unwanted features and attributes are removed: file links, emphasises, comments, indents, HTML, references etc.

\begin{table*}[h!]
    \centering
    \scalebox{0.85}{
    \begin{tabular}{|c|c|c|c|c|c|c|c|c|}
        \hline
        \multirow{2}{*}{Model} & \multirow{2}{*}{Tokens} & \multirow{2}{*}{Tokenizer} & Ratio & \multicolumn{5}{c|}{Benchmark Performance} \\
        \cline{4-9}
         & & & WMT:Wiki & Av. & HellaSwag & TruthfulQA & PIQA & WSC \\
        \hline
        GPT-2 & 1.5B & BPE & -:- & 0.4199 & 0.3114 & 0.4069 & 0.6251 & 0.4327 \\
        \hline
        \multirow{3}{*}{Dia\_2020} & 2.5B & BPE & 0.4:0.6 & 0.4820 & 0.2545 & 0.5289 & 0.5098 & 0.6346 \\
                                   & 5B   & BPE & 0.4:0.6 & 0.4817 & 0.2573 & 0.5251 & 0.5098 & 0.6346  \\
                                   & 10B  & BPE & 0.4:0.6 & 0.4639 & 0.2560 & 0.5022 & 0.5299 & 0.5673 \\
        \hline
        \multirow{2}{*}{Dia\_2020} & 2.5B & BPE & 0.4:0.6 & 0.4820 & 0.2545 & 0.5289 & 0.5098 & 0.6346 \\
                                   & 2.5B & SP & 0.4:0.6 & 0.4773 & 0.2550 & 0.5255 & 0.5131 & 0.6154 \\  
        \hline
        \multirow{7}{*}{2011} & 2.5B & BPE & 0.2:0.8 & 0.4787 & 0.2535 & 0.5227 & 0.5038 & 0.6346 \\
                              & 2.5B & BPE & 0.3:0.7 & 0.4785 & 0.2517 & 0.5266 & 0.5011 & 0.6346 \\
                              & 2.5B & BPE & 0.4:0.6 & 0.4792 & 0.2525 & 0.5259 & 0.5038 & 0.6346 \\
                              & 2.5B & BPE & 0.5:0.5 & 0.4794 & 0.2517 & 0.5262 & 0.5049 & 0.6346 \\
                              & 2.5B & BPE & 0.6:0.4 & \textbf{0.4808} & 0.2514 & 0.5283 & 0.5087 & 0.6346 \\
                              & 2.5B & BPE & 0.7:0.3 & \underline{0.4808} & 0.2489 & 0.5269 & 0.5126 & 0.6346 \\
                              & 2.5B & BPE & 0.8:0.2 & 0.4788 & 0.2489 & 0.5262 & 0.5049 & 0.6346 \\
        \hline
        \multirow{7}{*}{2020} & 2.5B & BPE & 0.2:0.8 & 0.4786 & 0.2522 & 0.5233 & 0.5044 & 0.6346 \\
                              & 2.5B & BPE & 0.3:0.7 & 0.4798 & 0.2526 & 0.5242 & 0.5076 & 0.6346 \\
                              & 2.5B & BPE & 0.4:0.6 & \textbf{0.4820} & 0.2545 & 0.5289 & 0.5098 & 0.6346 \\
                              & 2.5B & BPE & 0.5:0.5 & 0.4781 & 0.2513 & 0.5212 & 0.5054 & 0.6346 \\
                              & 2.5B & BPE & 0.6:0.4 & \underline{0.4805} & 0.2507 & 0.5288 & 0.5077 & 0.6346 \\
                              & 2.5B & BPE & 0.7:0.3 & 0.4799 & 0.2509 & 0.5193 & 0.5147 & 0.6346 \\
                              & 2.5B & BPE & 0.8:0.2 & 0.4795 & 0.2505 & 0.5220 & 0.5109 & 0.6346 \\
        \hline
    \end{tabular}}
    \caption{Performance comparison of different models trained, including GPT-2 for reference. Benchmarks: HellaSwag, TruthfulQA, PIQA, and WSC.}
    \label{table:model_comparison}
\end{table*}

\section{Benchmarks}
\label{app:bench_datasets}

\textbf{HellaSwag} \cite{zellers2019hellaswag} (10-shot, acc\_norm, 10,042 samples) - a commonsense inference task that has very high human performance (>95\%) yet challenges LLMs.
\\\\
\textbf{TruthfulQA} \cite{lin2022truthfulqa} (0-shot, mc2, 817 samples) - a task that measures whether models give truthful answers and do not reproduce human falsehoods.
\\\\
\textbf{PIQA} \cite{bisk2019piqa} (1-shot, acc\_norm, 1,838 samples) - a physical commonsense reasoning task designed to test models' knowledge of the real world. This is another dataset that humans find very easy (95\% accuracy). 
\\\\
\textbf{WSC} \cite{levesque2012winograd} (5-shot, acc, 273 samples) - a binary QA problem that requires world knowledge and reasoning skills.
\\\\
\textbf{Winogrande} \cite{sakaguchi2019winogrande} (5-shot, acc, 44,000 samples) - a larger, harder version of the WSC dataset \cite{levesque2012winograd}.

\section{Model Tuning}
\label{app:tuning}

The following section outlines the process for deciding which assumptions to make and parameters to use in the creation of our training datasets and models. We used the 2020 for the majority of the tuning and tested the tokenizer, dataset size, and data domain split ratio. The tuning was not rigorous since training every configuration of the models would have been computationally prohibitive and unproductive. 

\subsection{Tokenizer}
\label{app:tuning_tokenizer}

\cite{radford2019language} used a BPE tokenizer to originally train GPT-2. However there have been many papers that have shown that BPE is problematic in the way it segments words \cite{hofmann-etal-2021-superbizarre}. As a result, we tested the BPE tokenizer against a Sentence Piece tokenizer. The search for the optimal tokenizer was far from extensive, but from the two tokenizers BPE performed better so it was selected to train the rest of the models.

\subsection{Dataset Size}
\label{app:tuning_data_size}

Although \cite{hoffmann2022training} showed that the ratio of tokens to parameters should be 20:1 for complete pre-training, we wanted to test the effect of adding more data than the required amount. Therefore we tested the performance of using a 5B and 10B token training dataset and ran the training for 1 epoch. The datasets were constructed in exactly the same way as the 2.5B token dataset and were just sampled for longer until the required token count was met. Table \ref{table:model_comparison} shows clearly that dataset size does not effect the downstream performance on our benchmark datasets.

\subsection{Domain Split}
\label{app:tuning_domain_split}

We also fine-tuned the proportion of each data domain within the training dataset. Previous research, as noted in Section \ref{sec:data_agg}, has shown that the type of domain in the training data can influence downstream performance. Therefore, we determined the optimal proportion of each dataset that yielded the best results for both the 2011 and 2020 data. The comparison between two models at different extremes of our time period meant that we could feel more confident that the optimal ratio split was consistent across time. Given that the 0.6:0.4 ratio of WMT to Wiki data was the best performing in 2011 and the second best in 2020 we went with this domain split for all of our models.

\section{Dataset Histogram}
\label{app:data_hist}

\begin{figure}[h]
    \centering
    \includegraphics[width=\linewidth]{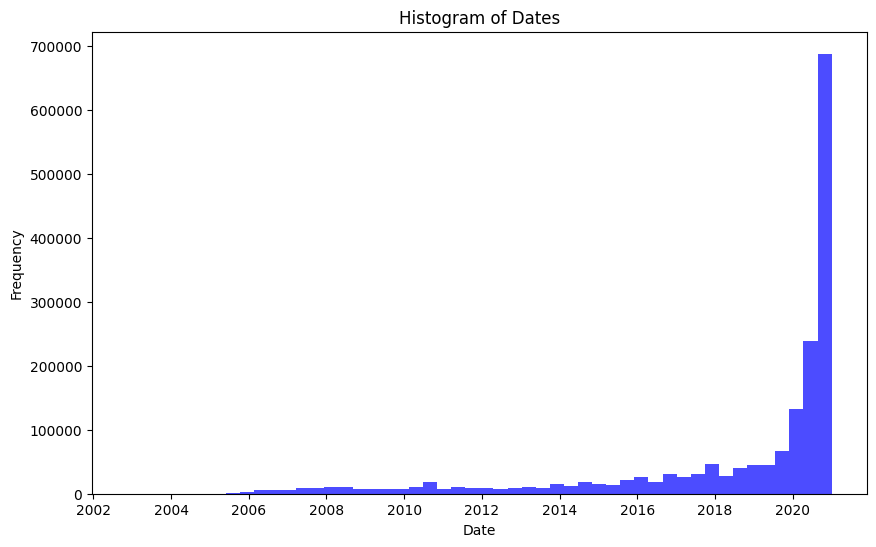}
    \caption{Histogram of the publication of each Wikipedia article revision in the 2020 training dataset.}
    \label{fig:wiki_hist}
\end{figure}

The two base datasets, WMT and Wikipedia, used to create the training dataset used different timestamp formats. For Wikipedia, the most recent revision to the cut off date was the version used in creating the yearly datasets. This meant that although Wikipedia was treated as a snapshot in time and was accordingly randomly sampled, some revision versions were older than others. The histogram in Figure \ref{fig:wiki_hist} outlines the revision publication dates of each of the samples in the 2020 dataset, with the maximum date 2020-12-31. The WMT dataset only exists in yearly buckets, which limits the granularity of the dataset. There is no data used past the cutoff date but the exact distribution across the months and weeks is not possible to know.

\section{Perplexity Calculation}
\label{app:perp}

In our perplexity calculations, we deviate from the conventional methodology of computing perplexity (PPL) of a language model, where some preceding context is usually considered. Instead, we calculate the PPL with zero context.

Formally, the perplexity of a sequence \( X = (x_0, x_1, \ldots, x_t) \) without considering any preceding tokens is given by:
\begin{equation}
    \text{PPL}(X) = \exp\left\{-\frac{1}{t} \sum_{i=0}^{t} \log p(x_i)\right\}
\end{equation}
where \( p(x_i) \) is the model's estimated probability of the token \( x_i \), independent of any preceding sequence.

This zero-context perplexity enables us to understand the models' comprehension of an individual word, without being biased by the context that precedes it.

\begin{figure}[h]
    \centering
    \includegraphics[width=\linewidth]{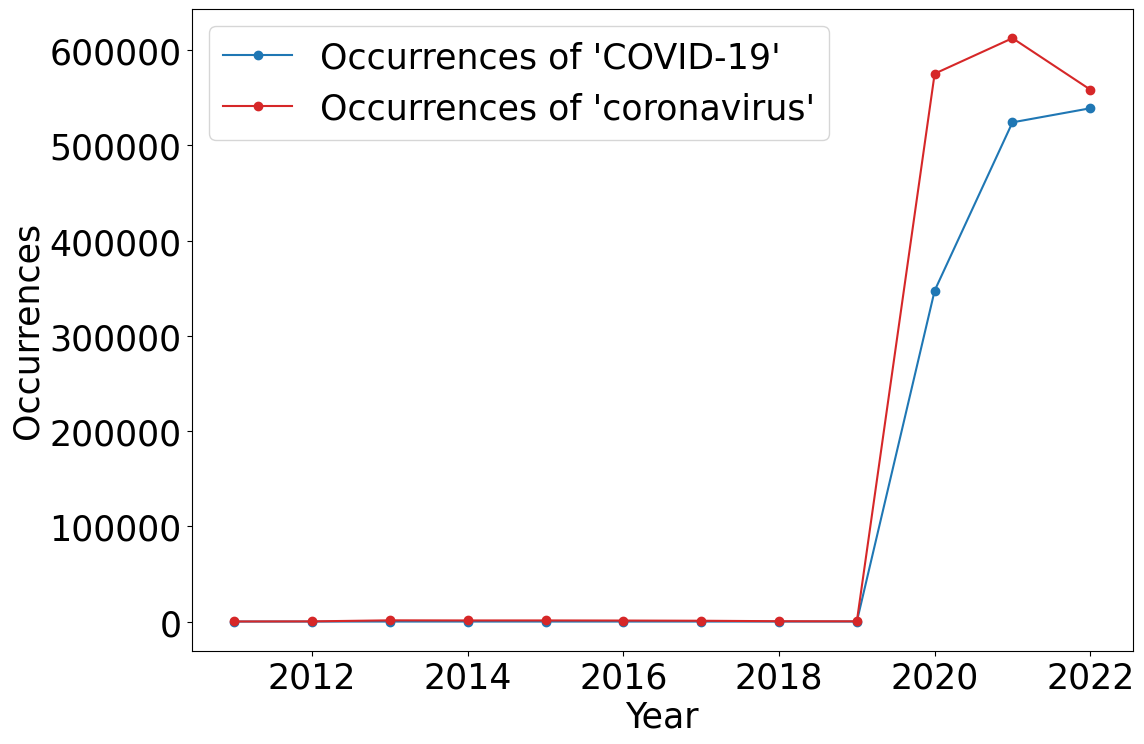}
    \caption{Number of occurrences of the words \textcolor{red}{coronavirus} and \textcolor{blue}{COVID-19} in the training datasets.}
    \label{fig:training_data}
\end{figure}

To visualise the effect that the training data has on the model, Figure \ref{fig:covid_perp} shows the perplexity of the words "COVID-19" and "coronavirus" using the TiMaGPT models. We would expect the model to have no real knowledge of what COVID-19 is before 2020 and then a significant understanding during and after. Figure \ref{fig:covid_perp} shows that this is the case for TiMaGPT models, as the models all have very high perplexity before the pandemic and very low perplexity after. This is due to the differences in the training datasets. Figure \ref{fig:training_data} shows the different exposures the models had to the words "COVID-19" and "coronavirus". 

\section{Licenses}
\label{app:licenses}

\subsection{External Licenses}
\label{app:ext_licenses}

In the making of our training datasets and training of our models we used data and code that were licensed in ways that might be of interest to the reader. The \textbf{Wikipedia} dump data is licensed under a \textit{GNU Free Documentation License (GFDL)} and the \textit{Creative Commons Attribution-Share-Alike 3.0 License}, two very permissive licenses. The \textbf{WMT News} data is released under the same terms as the ParaCrawl dataset \footnote{https://www.paracrawl.eu}, meaning that WMT claim no ownership over the text and that the packaging of the data is released under a \textit{Creative Commons CC0 Licence}, which means that they do not reserve any rights over the way the data is assembled. There is however some copyrighted material in the dataset, which we use under Fair Use \footnote{Fair Use (US): http://tinyurl.com/497jze9m} and Fair Dealing \footnote{Fair Dealing (UK): http://tinyurl.com/5f7nw4tu} principles. 

We have also used various software packages when creating these models, which can all be accessed under permissive licenses. The \textit{lm-eval-harness} package, which was used to evaluate the models, is released under an MIT License \footnote{http://tinyurl.com/bdeapaze}. The \textit{transformers} and \textit{optimum-graphcore} packages, which were used to train the models, are released under and Apache 2.0 License \footnote{http://tinyurl.com/yc7mvkny} \footnote{http://tinyurl.com/mspzm9jp}. 

\subsection{Our Licenses}
\label{app:int_licenses}

We release our models and datasets in accordance with the licenses of the original works. We do not claim ownership over any of the material used. We license the packaging of the data and models under a Creative Commons CC0 license ("no rights reserved"). The datasets for the models are of academic interest and therefore fall under Fair Use/Dealing principles. However we will comply with any legal requests pertaining to our data if we are legally compelled to do so.

\section{Emissions}
\label{app:emissions}

Training models has both computational and environmental implications. The energy consumption of training large language models can be substantial. To quantify this, we calculated the energy consumption of training all of our models and the associated carbon emissions. The computational costs for cleaning the datasets are not considered but are significant: the Wikipedia datasets took several days to extract and clean. The computational costs for evaluation are also not considered but are significant: each model was evaluated extensively. 

Our models were trained using a GraphCore Pod with 16 IPU-M2000 chips, which each consumes a maximum of 6kW of power \cite{graphcore2022ipupod16}. To train all of the models in this paper the POD-16 was consumed 388.40 kWh.

The IPU POD-16 is situated in Charlotte, North Carolina. Given the carbon emissions from this grid is $328 \text{gCO}_2\text{eq/kWh}$ \footnote{http://tinyurl.com/2bnsv8yh}, the carbon emissions associated with the energy used for a single model training can be deduced:
\begin{align}
    \text{Emissions} &= E \times \text{Carbon Intensity} \\
    &= 388.40\text{kWh} \times 342 \text{gCO}_2\text{eq/kWh} \\
    &= 132,832.80 \text{gCO}_2\text{eq}
\end{align}
or equivalently, $132.83\text{kgCO}_2\text{eq}$.

Whilst this is significant, the emissions are significantly reduced by the hardware that the models were trained on. The Graphcore POD-16 is very efficient which means that the emissions associated with the training of the models are less than the average transatlantic flight \cite{mazareanu2020co2}.

\end{document}